\documentclass[conference]{IEEEtran}
\IEEEoverridecommandlockouts
\usepackage{cite}
\usepackage{amsmath,amssymb,amsfonts}
\usepackage{algorithmic}
\usepackage{graphicx}
\usepackage{textcomp}
\usepackage{xcolor}
\def\BibTeX{{\rm B\kern-.05em{\sc i\kern-.025em b}\kern-.08em
    T\kern-.1667em\lower.7ex\hbox{E}\kern-.125emX}}
\begin{document}

\title{Hate Speech Detection on Vietnamese Social Media Text using the Bidirectional-LSTM Model
}

\author{\textbf{Hang Thi-Thuy Do}, \textbf{Huy Duc Huynh}, \textbf{Kiet Van Nguyen}, \textbf{Ngan Luu-Thuy Nguyen} and \textbf{Anh Gia-Tuan Nguyen}\\ University of Information Technology, VNU-HCM\\{\tt\{16520339, 16520508\}@gm.uit.edu.vn, \{kietnv, ngannlt, anhngt\}@uit.edu.vn}}

\maketitle

\begin{abstract}
In this paper, we describe our system which participates in the shared task of Hate Speech Detection on Social Networks of VLSP 2019 evaluation campaign. We are provided with the pre-labeled dataset and an unlabeled dataset for social media comments or posts. Our mission is to pre-process and build machine learning models to classify comments/posts. In this report, we use Bidirectional Long Short-Term Memory to build the model that can predict labels for social media text according to Clean, Offensive, Hate. With this system, we achieve comparative results with 71.43\% on the public standard test set of VLSP 2019.

\end{abstract}


\section{Introduction}
In recent years, social networking has grown and become prevalent with every people, it makes  easy for people to interact and share with each other. However, every problem has two sides. It also has some negative issues, hate speech is a hot topic in the domain of social media. With the freedom of speech on social networks and anonymity on the internet, some people are free to comment on hate and insults. Hate speech can have an adverse effect on human behavior as well as directly affect society. We don't manually delete each of those comments, which is time-consuming and boring. This spurs research to build an automated system that detects hate speech and eliminates them. With that system, we can detect and eliminate hate speech and thus reduce their spread on social media. With Vietnamese, we can use methods to apply specific extraction techniques manually and in combination with string labeling algorithms such as Conditional Random Field (CRF)[1], Model Hidden Markov (HMM)[2] or Entropy[3]. However, we have to choose the features manually to bring the model with high accuracy. Deep Neural Network architectures can handle the weaknesses of the above methods. In this report we apply Bidirectional Long Short-Term Memory (Bi-LSTM) to build the model. Also combined with the word embedding matrix to increase the accuracy of the model. 

The rest of the paper is organized as follows. In section 2, we presented the related work. In section 3, we described our Bi-LSTM system. In sections 4 and 5, we presented the experimental process and results. Finally, section 6 gives conclusions about the work. 

\section{Related Work}
Gao and Huang (2017)[4] used BiLSTMs with attention mechanism 372 to detect hate speech. They illustrated that the Bi-directional LSTM model with attention mechanism achieves the high performance. They hypothesize that this is because hate indicator phrases are often concentrated in a small region of a comment, which is especially the case for long comments. Davidson et al. (2017)[5] train a model to differentiate among three classes: containing hate speech, only offensive language, or neither.Jing Qian, Mai ElSherief, Elizabeth Belding, William Yang Wang (2018) [6] worked on classifying a tweet as racist, sexist or neither by  multiple deep learning architectures. ABARUAH at SemEval-2019 [7] presented the results obtained using bi-directional long short-term memory (BiLSTM) with and without attention and Logistic Regression (LR) models for multilingual detection of hate speech against immigrants and women in Twitter. Animesh Koratana and Kevin Hu [8] use many machine learning models to detect toxic words, in which the Bi-Lstm model got the highest performance. Malmasi and Zampieri (2017)[9] made a similar study to compare the performance of different features in detecting hate speech.

\section{Bi-LSTM model for Vietnamese Hate Speech Detection}
As mentioned previously, we propose a framework based on the ensemble of Bi-LSTM models to perform hate speech detection with the provided dataset. Besides, we also implemented some more models to compare and find the optimal model for the task.

\subsection{Long Short-Term Memory}

LSTM takes words from an input sentence in a distributed word representation format. LSTM's network architecture includes memory cells and ports that allow the storage or retrieval of information. These gates help the LSTM memory cell to perform a write, read and reset operation. They enable the LSTM memory cell to store and access information over a period of time. 

\subsection{Bidirectional Long Short-Term Memory}

One drawback of LSTM architecture[10] is that they are
only considering the previous context. However, the identification of a word depends not only on the previous context but also on the subsequent context. Bidirectional LSTM (Bi-LSTM)[11] was created to overcome this weakness. A Bi-LSTM architecture usually contains two single LSTM networks used simultaneously and independently to model input chains in two directions: forward LSTM and backward LSTM.

\begin{figure}[htbp]
\centerline{\includegraphics{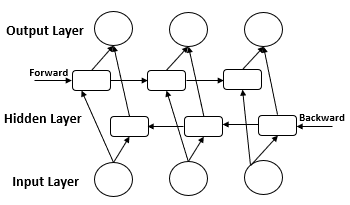}}
\caption{Bi-LSTM architecture [12]}
\label{fig}
\end{figure}

\section{Pre-Processing}
The pre-processing performed on the text includes
the following:
\begin{itemize}
\item The comments were converted to lowercase.
\item The URLs, mentions (@) and non-alphabetic characters are removed (number, excess whitespace).
\item Several stopwords were removed from the comments. We don't remove all stopword because having a few stopwords affect the results.
\item A few characters that don't affect the results are replaced by an empty string. 
\item Using Tokenizer to convert each comment into a sequence of integers.
\end{itemize}

\section{Experiments}
\subsection{Dataset and Word Embeddings}
VLSP Shared Task 2019: Hate Speech Detection on Social Networks: This dataset includes 25431 items in csv format, the dataset was divided into two file, training dataset with 20345 items and test dataset with 5086 items. Each data line of training dataset is assigned 1 of 3 labels CLEAN, OFFENSIVE or HATE. The test dataset is not assign label. The statistic summarization of the given training dataset is described in Table I.

\begin{itemize}
\item \textbf{Hate speech (HATE)} contains the abusive language, which often bears the purpose of insulting individuals or groups, and can include hate speech, derogatory and offensive language. An item is identified as hate speech if it (1) targets individuals or groups on the basis of their characteristics; (2) demonstrates a clear intention to incite harm, or to promote hatred; (3) may or may not use offensive or profane words.
\item \textbf{Offensive but not hate speech (OFFENSIVE)} is an item (posts/comments) may contain offensive words but it does not target individuals or groups on the basis of their characteristics.
\item \textbf{Neither offensive nor hate speech (CLEAN)} is a normal item. It's conversations, expressing emotions normally. It does not contain offensive language or hate speech. 
\end{itemize}

In this paper, we use two different word embeddings to compare and find out the best word embedding such as Word2Vec [17] and FastText [16]. We used pre-trained vector with large dimensions to increase the accuracy of the model. Through experiments we found FastText achieved better results.

\begin{table}[htbp]
\caption{THE STATISTIC OF VLSP 2019 HSDoSN TRAINING DATASET}
\begin{center}
\begin{tabular}{|c|c|c|c|c|}
\hline

\cline{2-4} 
\textbf{} & \textbf{CLEAN} & \textbf{\textit{OFFENSIVE}}& \textbf{\textit{HATE}}& \textbf{\textit{TOTAL}} \\
\hline
Frequency & 18614& 1022 & 709 & 20345\\
\hline
Percentage & 91.49\%& 5.02\% & 3.49\% & 100\%\\
\hline

\end{tabular}
\label{tab1}
\end{center}
\end{table}

For this public dataset, we find that the dataset is an unbalanced dataset. The CLEAN label has the highest rate with 91.49\% and the HATE label is lowest with 3.49\%. Therefore, it is difficult and challenging to find a good model for this task. 

\subsection{Evaluation on each Model}
For problems of this type, there are many models suitable to handle such as: SVM, Bi-LTSM, LR, GRU, CNN and etc. To solve this problem, we implement four different models (SVM, LR, Bi-LSTM, and GRU) to compare and find the most suitable one. To evaluate the four models on this task, we divide the training dataset into two parts training, testing rate of 80\%, 20\% respectively. 

The details of our models are provided below.

\textbf{1. Support Vector Machine (SVM)}

Support Vector Machines (SVMs) are a popular machine learning method for classification, regression, and other learning tasks [13]. It is often used for two-class classification problems. For this problem, it has three labels, so we use the SVM to classify twice, two label at a time.Firstly, we classify two label 0 and 1, we achieved accuracy, precision, recall, and F1-score rates of 96.00\%, 93.37\%, 98.96\%, and 96.08\% respectively, on training dataset. Second time, we classify two label 1 and 2, we achieved accuracy, precision, recall, and F1-score rates of 84.34\%, 87.38\%, 78.86\%, and 82.90\% respectively. We find that this model doesn't classify well for two labels 1 and 2. Moreover, when we check this model with the public-test, it brings the result as not good as we expected with 63.87\%.

\textbf{2. Logistic Regression (LR)}

Logistic regression is basically a supervised classification algorithm. In a classification problem, the target variable(or output), can take only discrete values for a given set of features(or inputs) [14]. We have applied it to this problem as follows: Firstly, we use the TfidfVectorizer tool to convert text into feature vectors that are used as input for the model. Then, we used the Logistic Regression model to predict the classification results. When checking it on training datasets, we achieved accuracy, precision, recall, and F1-score rates of 94.17\%, 88.87\%, 55.54\%, and 64.15\% respectively. We also try submitting this model's result on the system, the result is worse we thought with 51.15\%

\textbf{3. Gated Recurrent Units (GRU)}

The Recurrent Neural Network (RNN) handles the variable-length sequence by having a recurrent hidden state whose activation at each time is dependent on that of the previous time [15]. The GRU is a variant of RNN and it only has two inputs. We have used it into this problem as follows: We have used it with word embeddings Fasttext [16]. First, we use Tokenizer() for sequences because GRU is good at processing long sequences. Then, we have applied this model to the problem. We achieved accuracy, precision, recall, and F1-score rates of 94.61\%, 67.12\%, 59.66\%, and 64.15\% respectively, on the training dataset. When we check this model with the public dataset, it brings the result quite good with 65.01\%

\textbf{4. Bidirectional Long Short-Term Memory (Bi-LSTM)}

The LSTM is a famous variant of RNN. The Bidirectional Long Short Term Memory can be trained using all available input information in the past and future of a specific time frame. We have applied it with word embeddings Fasttext [16] and baomoi.vn.model.txt [17]. As follows:

As well as the GRU, we also used GloVe Embedding for sequences and applied this model for the problem. When using word embeddings baomoi.vn.model.txt [17], we achieved accuracy, precision, recall, and F1-score rates of 93.26\%, 90.74\%, 50.30\%, and 53.62\% respectively, on the training dataset. The same with word embeddings Fasttext [16], we achieved accuracy, precision, recall, and F1-score rates of 95.67\%, 85.61\%, 67.36\%, and 73.84\% respectively, on the training dataset. We find that when combining the Bi-LSTM with fasttext will bring the result better. When we check it with the public dataset, it brings the result good with 71.43\%

\subsection{Experimental Results}
After conducting experiments on many models, we obtained the following results on public-test, shown in Table \ref{tab:ex}.

\begin{table}[ht]
\centering
\caption{\label{tab:ex}The results table of models.}
\begin{tabular}{l|r}
\textbf{Model} & \textbf{F1-Score} \\\hline
SVM & 63.87 \\
LR & 51.15 \\
GRU & 65.01\\
\textbf{Bi-LSTM} & \textbf{71.43}\\

\end{tabular}

\end{table}

We achieved the best result with Bi-LSTM, ranking the 2nd of the scoreboard on the public-test set shown in Table \ref{tab:rank}. However, our result ranks the 6th of the scoreboard on the private-test set.

\begin{table}[ht]
\centering
\caption{\label{tab:rank}The results table of the top 5 on public-test set}
\begin{tabular}{l|l|r}

\textbf{Rank} & \textbf{Team} & \textbf{F1-score} \\ \hline
1 & Try hard & 73.01 \\
\textbf{2} & \textbf{HH\_UIT} & \textbf{71.43}\\
3 & titanic & 70.74\\
4 & ABCD & 70.58\\
5 & TIN HUYNH & 70.57\\

\end{tabular}
\end{table}

\section{Conclusion and Future Work}
In this paper, we have presented our approach to address Vietnamese hate speech detection task proposed at the VLSP Shared Task 2019. We develop the system using Bidirectional Long Short Memory for classifying three different labels in this task. We participate in this and evaluate the performance of our system on this dataset. As a result, our result is 71.43\% of F1-score, ranking the 2nd of the scoreboard on the public-test set. \newline

In the future work, we plan to address this problem in different ways to enhance the performance of this task. We will investigate experiments both in traditional machine learning and types of deep learning for this problem. In addition, we also analyze experimental results on this task to choose the efficient approach such as the hybrid approach which combines machine learning and rule-based approaches to boost the result of detecting hate speech on Vietnamese social media text.

\section*{Acknowledgment}
We would like to thank the VLSP Shared Task 2019 organizers for their really hard work and providing the dataset of Vietnamese Hate Speech Detection on social networks for our experiments.

\end{document}